\documentclass[11pt,a4paper]{article}
\usepackage{times}
\usepackage{url}
\usepackage{latexsym}
\usepackage{graphicx}
\usepackage[utf8]{inputenc}
\usepackage[T1]{fontenc}
\usepackage[plain]{fancyref}
\usepackage{eacl2017}

\newcommand*{\langiso}[2]{#1 [\emph{#2}]}

\eaclfinalcopy 
\def\eaclpaperid{101} 

\title{Continuous multilinguality with language vectors}

\author{Robert Östling \\
  Department of Linguistics\thanks{Work done while the author was at the
  University of Helsinki} \\
  Stockholm University \\
  {\tt robert@ling.su.se} \\\And
  Jörg Tiedemann \\
  Department of Modern Languages \\
  University of Helsinki \\
  {\tt jorg.tiedemann@helsinki.fi} \\}

\date{}

\begin{document}
\maketitle
\begin{abstract}
    Most existing models for multilingual natural language processing (NLP)
    treat language as a discrete category, and make predictions for either
    one language or the other. In contrast, we propose using continuous vector
    representations of language. We show that these can be learned efficiently
    with a character-based neural language model, and used to improve
    inference about language varieties not seen during training.
    In experiments with 1303 Bible translations into 990 different languages,
    we empirically explore the capacity of multilingual language models,
    and also show that the language vectors capture genetic
    relationships between languages.
\end{abstract}

\section{Introduction}

Neural language models
\cite{Bengio2003aneural,Mikolov2010recurrent,Sundermeyer2012lstm} have
become an essential component in several areas of natural language processing
(NLP), such as machine translation, speech recognition and image captioning.
They have also become a common benchmarking application in machine learning
research on recurrent neural networks (RNN), because producing an accurate
probabilistic model of human language is a very challenging task which
requires all levels of linguistic analysis, from pragmatics to phonology,
to be taken into account.

A typical language model is trained on text in a single language, and if
one needs to model multiple languages the standard solution is to train a
separate model for each language. This presupposes large quantities of
monolingual data in each of the languages that needs to be covered and each
model with its parameters is completely independent of any of the other models.

We propose instead to use a single model with real-valued vectors to indicate
the language used, and to train this model with a large number of languages.
We thus get a language model whose predictive distribution
$p(x_t|x_{1\dots t-1},l)$ is a continuous function of the language vector $l$,
a property that is trivially extended to other neural NLP models.
In this paper, we explore the ``language space'' containing these vectors, and
in particular explore what happens when we move beyond the points representing
the languages of the training corpus. 

The motivation of combining languages into one single model is at least
two-fold: First of all, languages are related and share many features and
properties, a fact that is ignored when using independent models.
The second motivation is data sparseness, an issue that heavily influences the
reliability of data-driven models. Resources are scarce for most languages in
the world (and also for most domains in otherwise well-supported languages),
which makes it hard to train reasonable parameters. By combining data from
many languages, we hope to mitigate this issue.

In contrast to related work, we focus on massively multilingual data sets to cover for the
first time a substantial amount of the linguistic diversity in the world in a
project related to data-driven language modeling. We do not presuppose any
prior knowledge about language similarities and evolution and let the model
discover relations on its own purely by looking at the data. The only
supervision that is giving during training is a language identifier as a
one-hot encoding. From that and the actual training examples, the system
learns dense vector representations for each language included in our data set
along with the character-level RNN parameters of the language model itself.

\section{Related Work}

Multilingual language models is not a new idea
\cite{Fugen2003efficienthandling}, the novelty of our work lies primarily in
the use of language vectors and the empirical evaluation using nearly a
thousand languages.

Concurrent with this work, \newcite{Johnson2016zeroshot} conducted a study
using neural machine translation (NMT), where a sub-word decoder is told which
language to generate by means of a special language
identifier token in the source sentence. This is close to our model, 
although beyond a simple interpolation experiment (as in our
\Fref{sec:generating}) they did not further explore the language vectors,
which would have been challenging to do given the small number of languages
used in their study.

\newcite{Ammar2016manylanguages} used one-hot language identifiers as input to
a multilingual word-based dependency parser, based on multilingual word
embeddings. Given that they report this resulting in higher accuracy than
using features from a typological database, it is a reasonable guess that
their system learned language vectors which were able to encode syntactic
properties relevant to the task. Unfortunately, they also did not look
closer at the language vector space, which would have been interesting given
the relatively large and diverse sample of languages represented in the
Universal Dependencies treebanks.


Our evaluation in \Fref{sec:clustering} calls to mind previous work on
automatic language classification, by \newcite{Wichmann2010evaluating} among
others. However, our purpose is not to detect genealogical relationships,
even though we use the strong correlation between such classifications and our
language vectors as evidence that the vector space captures sensible
information about languages.

\section{Data}

\begin{figure}
    \includegraphics[width=0.45\textwidth]{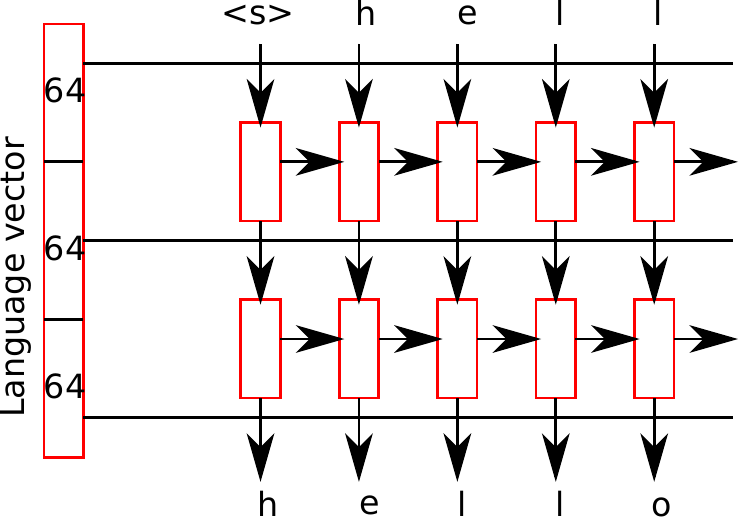}
    \caption{Schematic of our model. The three parts of the language vector
    are concatenated with the inputs to the two LSTM:s and the final
    softmax layer.}
    \label{fig:model}
\end{figure}

We base our experiments on a large collection of Bible translations crawled
from the web, coming
from various sources and periods of times. Any other multilingual data
collection would work as well, but with the selected corpus we have the
advantage that we cover the same genre and roughly the same coverage for
each language involved. It is also easy to divide the data into training and
test sets by using Bible verse numbers, which allows us to control for
semantic similarity between languages in a way that would have been difficult in
a corpus that is not multi-parallel.
Altogether we have 1,303 translations in 990 languages that we can use for
our purposes. These were chosen so that the model alphabet size is below 1000
symbols, which was satisfied by choosing only translations in Latin,
Cyrillic or Greek script.

Certainly, there are disadvantages as well, such as the limited size
(roughly 500 million tokens in total, with most languages having only one
translation of the New Testament each, with roughly 200 thousand tokens),
the narrow domain and the high overlap of named entities.
The latter can lead to some unexpected effects when using nonsensical language
vectors, as the model will then generate a sequence of random names.

The corpus deviates in some ways from an ideal multi-parallel corpus.
Most translations are of the complete New Testament, whereas around 300 also
contain the Old Testament (thus several times longer),
and around ten contain only portions of the New Testament.
Additionally, several languages have multiple translations, which
are then concatenated. These translations may vary in age and style, but
historical versions of languages (with their own ISO 639-3 code) are treated
as distinct languages.
During training we enforce a uniform distribution between languages when
selecting training examples.

\section{Methods}

Our model is based on a standard stacked character-based
LSTM \cite{Hochreiter1997lstm} with two layers, followed by a hidden layer
and a final output layer with softmax activations.
The only modification made to accommodate the fact that we train the model
with text in nearly a thousand languages, rather than one, is that language
embedding vectors are concatenated to the inputs of the LSTMs at each time
step and the hidden layer before the softmax. We used three separate embeddings
for these levels, in an attempt to capture different types of information
about languages.\footnote{The embeddings at the different levels are
different, but we did not see any systematic variation. We also found that
using the same embedding everywhere gives similar results.}
The model structure is summarized in \Fref{fig:model}.

In our experiments we use 1024-dimensional LSTMs, 128-dimensional character
embeddings, and 64-dimensional language embeddings.
Layer normalization \cite{Ba2016layernormalization} is used, but no dropout or
other regularization
since the amount of data is very large (about 3 billion characters) and
training examples are seen at most twice. For smaller models early stopping is
used. We use Adam \cite{Kingma2014adam} for optimization.
Training takes between an hour and a few days on a K40 GPU,
depending on the data size.


\section{Results}

In this section, we present several experiments with the model described.
For exploring the language vector space, we use 
hierarchical agglomerative clustering for visualization.
For measuring performance, we use cross-entropy on held out-data.
For this, we use a set of
the 128 most commonly translated Bible verses, to ensure that the held-out
set is as large and overlapping as possible among languages.

\subsection{Model capacity}

\begin{figure}
    \includegraphics[width=0.49\textwidth]{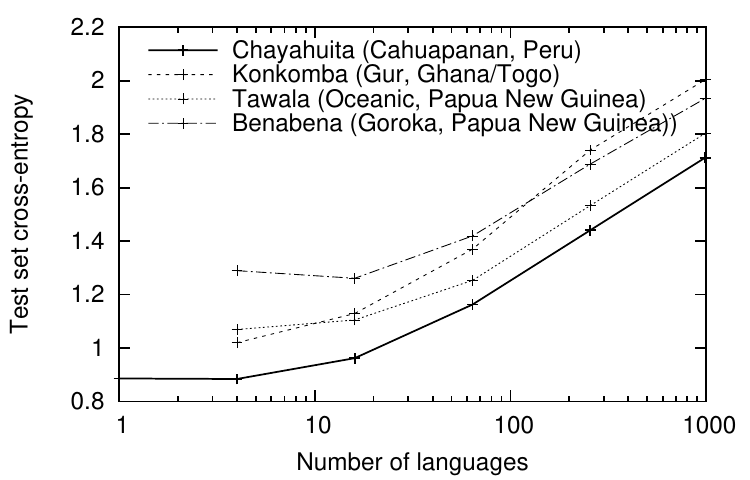}
    \caption{Cross-entropy of the test sets from the first four languages
    added to our model. At the leftmost point ($x = 1$),
    \emph{only} Chayahuita is used for training the model so no results
    are available for the other languages.}
    \label{fig:random}
\end{figure}

\begin{figure}
    \includegraphics[width=0.49\textwidth]{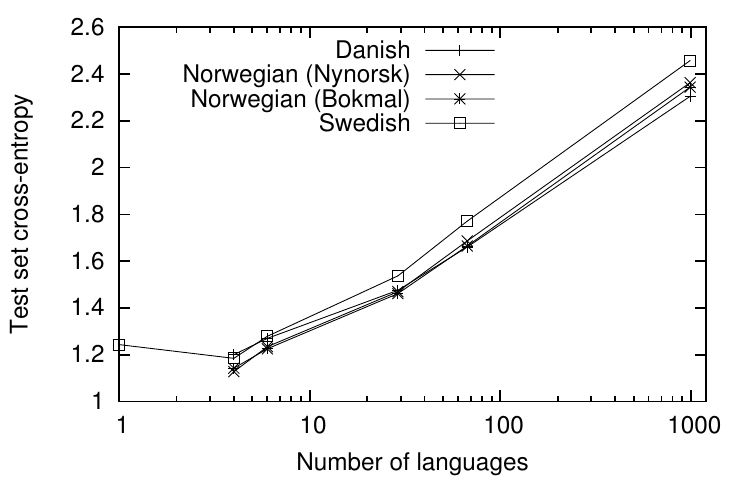}
    \caption{Cross-entropy of the test sets from Scandinavian languages.
    The languages added at each step are: Swedish, Norwegian+Danish,
    Icelandic+Faroese, remaining Germanic, remaining Indo-European,
    all remaining languages.}
    \label{fig:swe}
\end{figure}

Our first experiment tries to answer what happens when more and more languages
are added to the model. There are two settings: adding languages in a random
order, or adding the most closely related languages first.
Cross-entropy plots for these settings are shown in \Fref{fig:random} and
\Fref{fig:swe}.

\begin{figure}
    \includegraphics[width=0.49\textwidth]{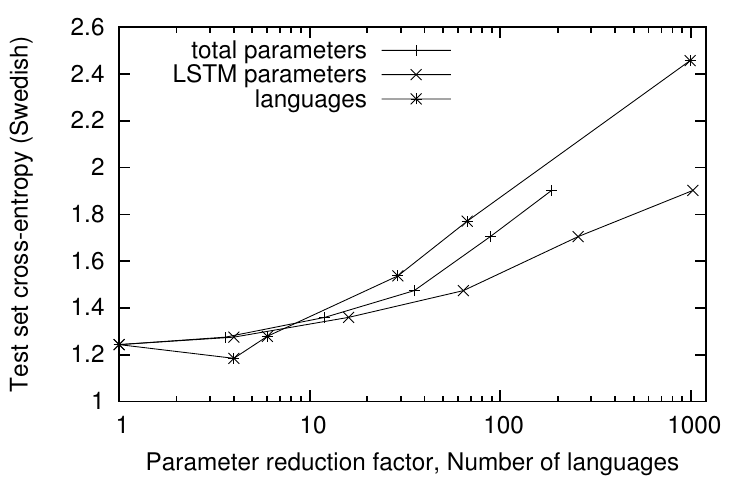}
    \caption{Cross-entropy of the Swedish test set, given two conditions:
    increasing number of languages by the given factor
    (adding the most similar languages first) or decreasing number of
    parameters by the same factor (for a monolingual model, which is why
    the curves meet at $x = 1$).
    }
    \label{fig:swesize}
\end{figure}

In both cases, the model degrades gracefully (or even improves) for a number
of languages, but then degrades linearly (i.e.\ exponential growth of
perplexity) with exponentially increasing number of languages.

For comparison, \Fref{fig:swesize} compares this to the effect of decreasing
the number of parameters in the LSTM by successively halving the hidden state
size.\footnote{Note that two curves are given, one counting \emph{all} model
parameters and one counting only the LSTM parameters. The latter dominates the
model size for large hidden states.}
Here the behavior is similar, but unlike
the Swedish model which got somewhat better when closely related languages
were added, the increase in cross-entropy is monotone. It would be interesting
to investigate how the number of model parameters needs to be scaled up in
order to accommodate the additional languages, but unfortunately the
computational resources for such an experiment increases with the number of
languages and would not be practical to carry out with our current equipment.

\subsection{Structure of the language space}
\label{sec:clustering}

\begin{figure}
    \includegraphics[width=1.05\columnwidth]{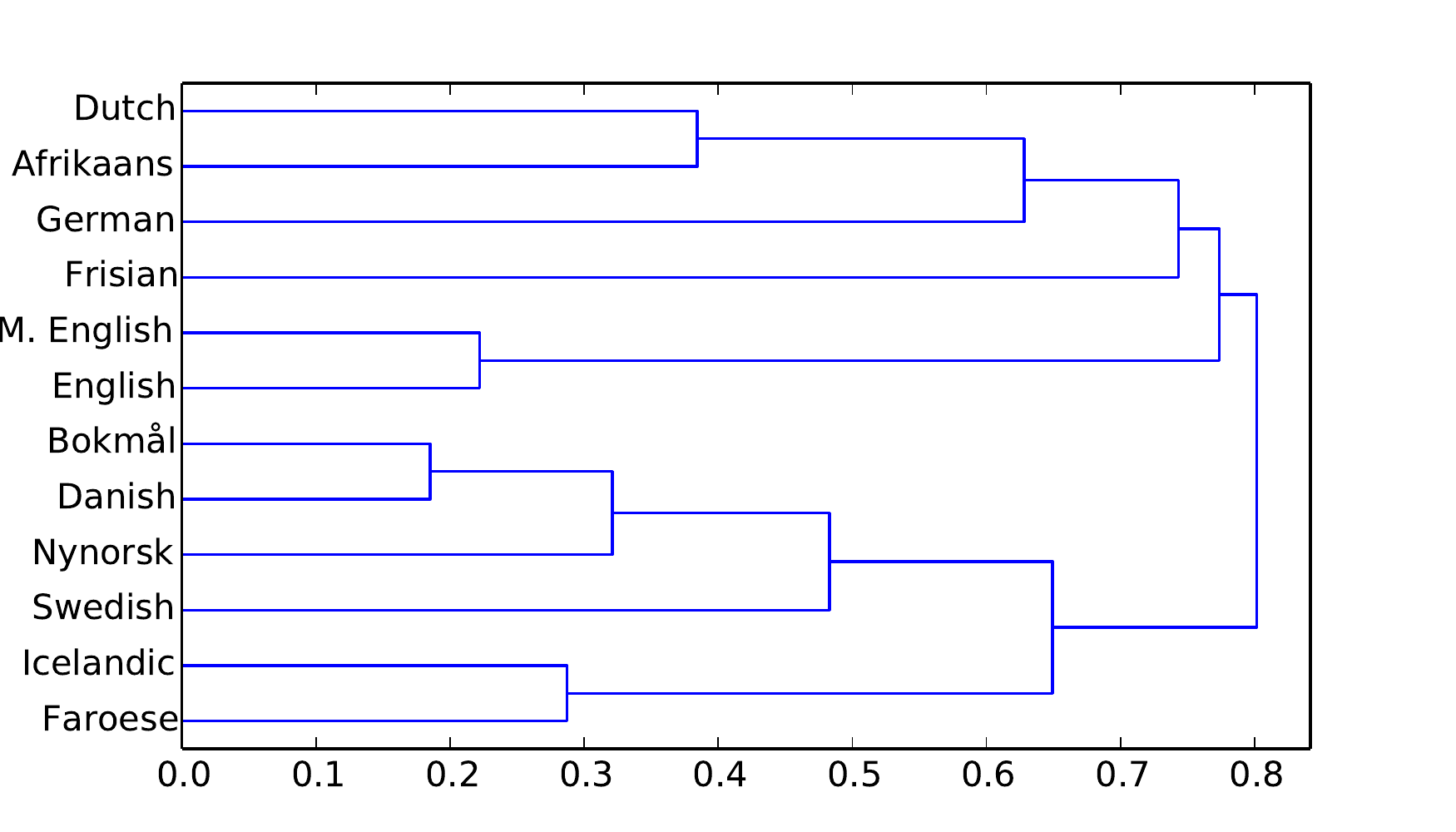}
    \caption{Hierarchical clustering of language vectors of Germanic languages.}
    \label{fig:germanic}
\end{figure}

We now take a look at the language vectors found during training with
the full model of 990 languages. \Fref{fig:germanic} shows a hierarchical
clustering of the subset of Germanic languages, which closely matches the
established genetic relationships in this language family.
While our experiments indicate that finding more remote relationships
(say, connecting the Germanic languages to the Celtic) is difficult for the
model, it is clear that the language vectors preserves similarity properties
between languages.

In additional experiments we found the overall structure of these clusterings
to be relatively stable across models, but for very similar languages (such as
Danish and the two varieties of Norwegian) the hierarchy might differ,
and the some holds for languages or groups that are significantly different
from the major groups. An example from \Fref{fig:germanic} is English,
which is traditionally classified as a West Germanic language with
strong influences from North Germanic as well as Romance languages. In the
figure English is (weakly) grouped with the West Germanic languages, but in
other experiments it is instead weakly grouped with North Germanic.

\subsection{Generating Text}
\label{sec:generating}

Since our language model is conditioned on a language vector, we can gain some
intuitive understanding of the language space by generating text from
different points in it. These points could be either one of the vectors
learned during training, or some arbitrary other point.
\Fref{tab:interpolation} shows text samples from different
points along the line between \langiso{Modern English}{eng} and
\langiso{Middle English}{enm}. Consistent with the results of
\newcite{Johnson2016zeroshot}, it appears that the interesting region lies
rather close to 0.5. Compare also to our \Fref{fig:eng-deu}, which shows that
up until about a third of the way between English and German,
the language model is nearly perfectly tuned to English.

\subsection{Mixing and Interpolating Between Languages}
\label{sec:interpolation}

\begin{figure}
    \includegraphics[width=1.0\columnwidth]{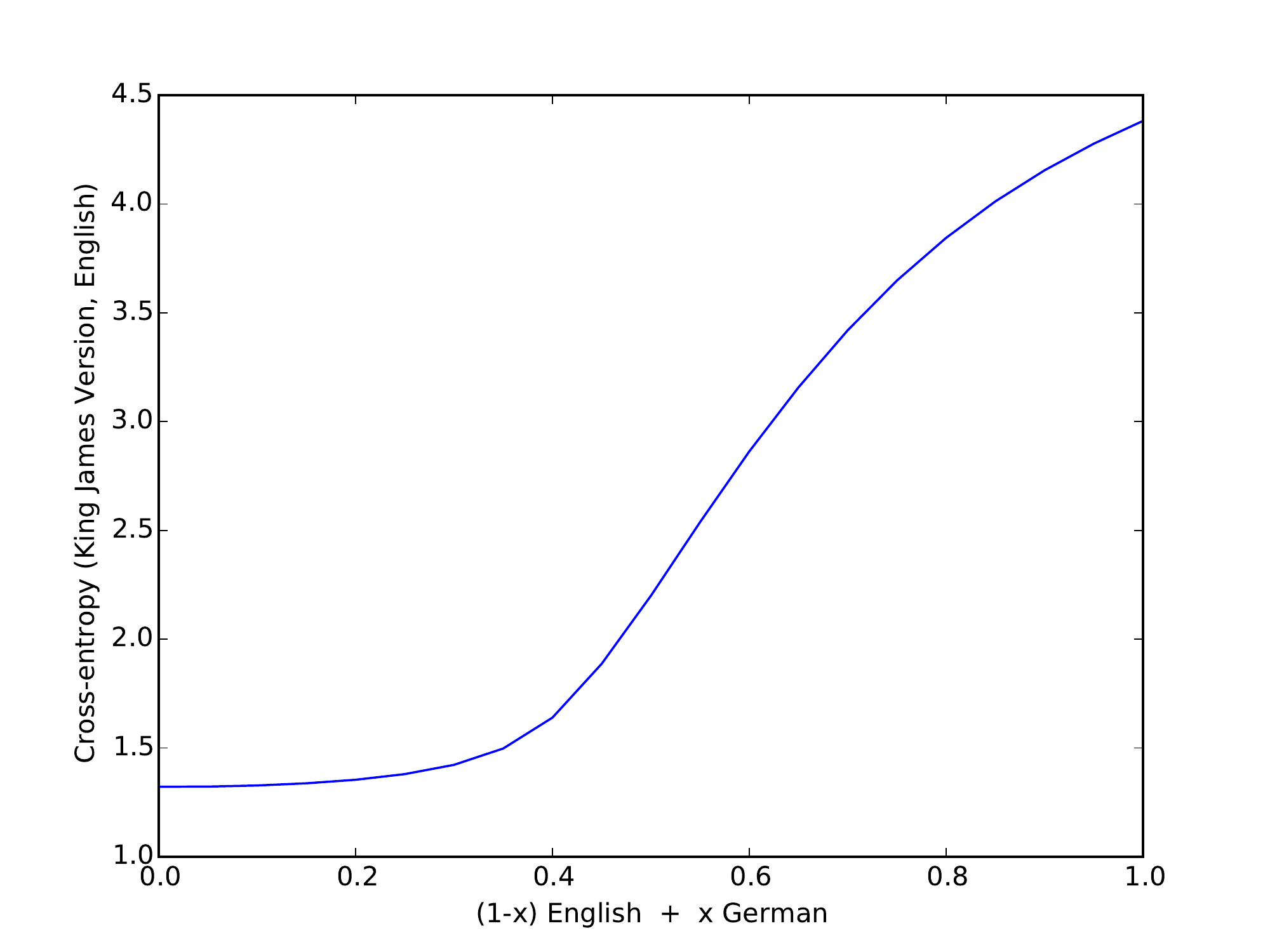}
    \caption{Cross-entropy of interpolated language models for English and
    German measured on English held-out text.}
    \label{fig:eng-deu}
\end{figure}

By means of cross-entropy, we can also visualize the relation between
languages in the multilingual space.  
Figure~\ref{fig:eng-deu} plots the interpolation results for two relatively
dissimilar languages, English and German.
As expected, once the language vector moves too close to the German one,
model performance drops drastically.

\begin{figure}
    \includegraphics[width=1.0\columnwidth]{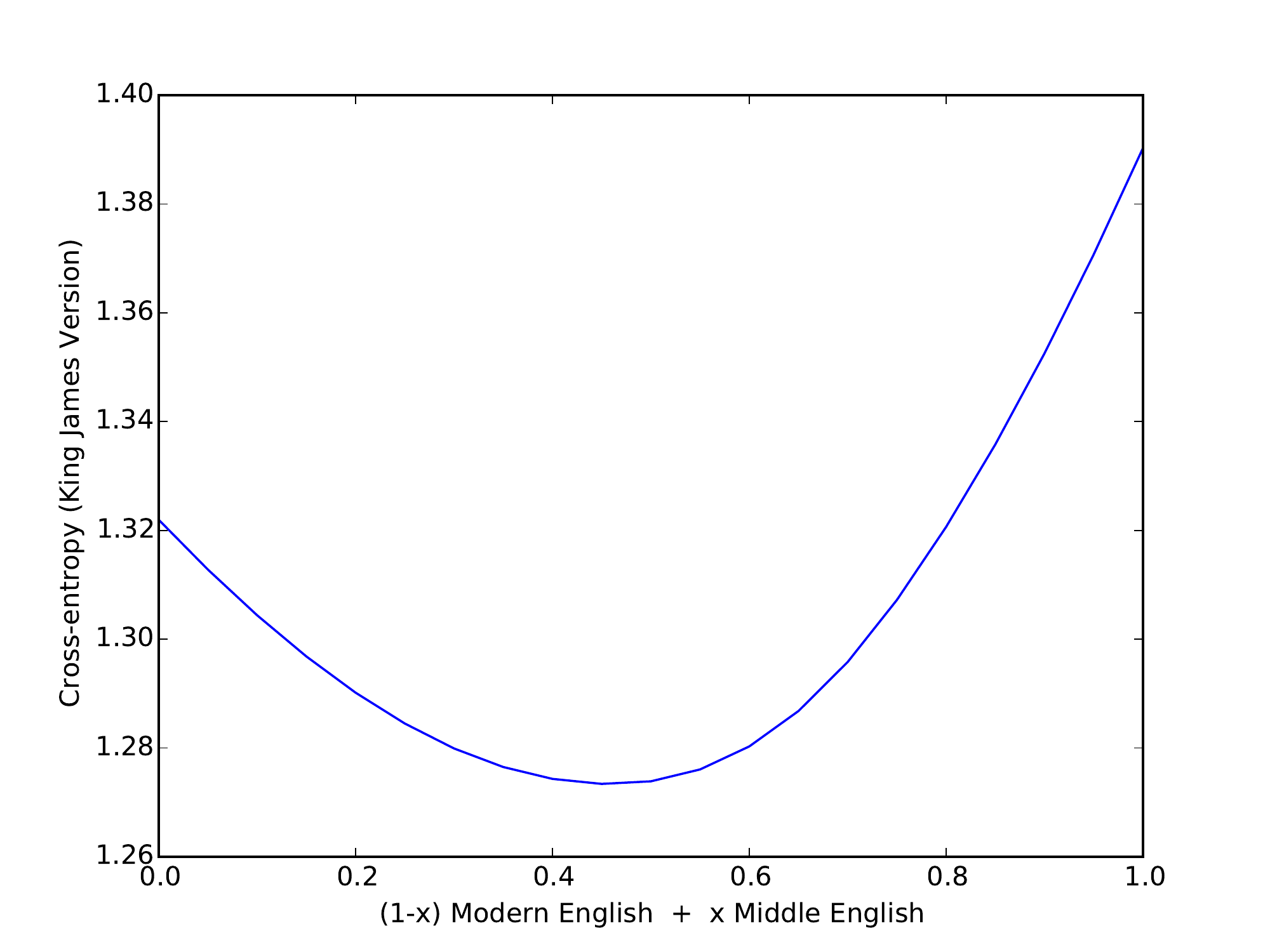}
    \caption{Cross-entropy of interpolated language models for modern and middle English tested on data from the King James Bible.}
    \label{fig:eng-enm}
\end{figure}

More interesting results can be obtained if we interpolate between two
language variants and compute cross-entropy of a text that represents an
intermediate form. \Fref{fig:eng-enm} shows the cross-entropy of the
King James Version of the Bible (published 1611), when interpolating between
Modern English (1500--) and Middle English (1050--1500).
The optimal point turns out to be close to the midway point between them.

\subsection{Language identification}

If we have a sample of an unknown language or language variant, it is possible
to estimate its language vector by backpropagating through the language model
with all parameters except the language vector fixed.\footnote{In practice,
    using error backpropagation is too computationally expensive for most
    applications, and we use it here because it requires only minimal
    modifications to our model. A more reasonable method could be to train a
    separate language vector encoder network.}
We found that a very small set of sentences is enough to give a considerable
improvement in cross-entropy on held-out sentences. In this experiment, we
used 32 sentences from the King James Version of the Bible. Using the
resulting language vector, test set cross-entropy improved from 1.39 (using
the Modern English language vector as initial value) to 1.35. This is
comparable to the result
obtained in \Fref{sec:interpolation}, except that here we do not restrict the
search space to points on a straight line between two language vectors.

\begin{table}
    \caption{Examples generated by interpolating between Modern English
    and Middle English.}
    \label{tab:interpolation}
    \vspace{6pt}
{\small
    \begin{tabular}{p{0.1\columnwidth}p{0.85\columnwidth}}
        \% & Random sample \newline (temperature parameter $\tau = 0.5$) \\
        \hline
        30 &
                and thei schulen go in to alle these thingis, and
                schalt endure bothe in the weie
            \\[6pt]
        40 &
                and there was a certaine other person who was
                called in a dreame that he went into a mountaine.
            \\[4pt]
        44 &
                and the second sacrifice, and the father, and the prophet,
                shall be given to it.
            \\[4pt]
        48 &
                and god sayd, i am the light of the world, and the powers of
                the enemies of the most high god may find first for many.
            \\[4pt]
        50 &
                but if there be some of the seruants, and to all the people,
                and the angels of god, and the prophets
            \\[4pt]
        52 &
                then he came to the gate of the city, and the bread was to be
                brought
            \\[4pt]
        56 &
                therefore, behold, i will lose the sound of my soul, and i
                will not fight it into the land of egypt
            \\[4pt]
        60 &
                and the man whom the son of man is born of god, so have i
                therefore already sent to the good news of christ.
     \end{tabular}
}
\end{table}

\section{Conclusions}

We have shown that \emph{language vectors}, dense vector representations of
natural languages, can be learned efficiently from raw text and possess several
interesting properties. First, they capture language similarity to the extent
that language family trees can be reconstructed by clustering the vectors.
Second, they allow us to interpolate between languages in a sensible way, and
even allow adopting the model using a very small set of text, simply by
optimizing the language vector.

%


\bibliography{eacl2017}
\bibliographystyle{eacl2017}

\end{document}